\def\year{2021}\relax
\documentclass[letterpaper]{article} 
\usepackage{aaai21}  
\usepackage{times}  
\usepackage{helvet,amsmath} 
\usepackage{courier,wrapfig}  
\usepackage[hyphens]{url}  
\usepackage{graphicx} 
\usepackage{amsmath,wrapfig, graphicx,bm}
\usepackage{amssymb}
\usepackage{natbib}  
\usepackage{caption} 
\graphicspath{{figures/}{Figures/}}
\newtheorem{definition}{Definition}

\newcommand{\tenq}[1]{\hbox{\oalign{${#1}$\crcr\hidewidth$\scriptscriptstyle\bm{\approx}$\hidewidth}}}
\newcommand\numberthis{\addtocounter{equation}{1}\tag{\theequation}}
\usepackage{algpseudocode}

\usepackage{paralist}
\usepackage{amsfonts}       

\usepackage{epsfig}
\usepackage{multicol}
\usepackage{vwcol}  
\usepackage{graphicx}
\graphicspath{{figures/}{Figures/}}
\usepackage{amsmath,wrapfig, graphicx,bm}
\usepackage{amssymb}

\usepackage{multirow}
\usepackage{colortbl, array}
\usepackage{hhline}
\usepackage[svgnames]{xcolor}
\usepackage{empheq}
\usepackage{booktabs}

\usepackage[utf8]{inputenc} 
\usepackage[T1]{fontenc}    
\usepackage{amsmath,amssymb,comment,tabularx}       
\usepackage{url}            
\usepackage{booktabs}       
\usepackage{nicefrac}       
\usepackage{microtype}      
\usepackage[linesnumbered,ruled]{algorithm2e}
\usepackage{graphicx,caption,color,wrapfig}
\usepackage{accents}
\usepackage{comment}
\usepackage[switch]{lineno}

\newcommand{\myrowcolour}{\rowcolor[gray]{0.925}}

\definecolor{rulecolor}{RGB}{70,10,171}
\definecolor{tableheadcolor}{RGB}{120,50,200}

\newcommand{\topline}{ %
        \arrayrulecolor{rulecolor}\specialrule{0.1em}{\abovetopsep}{0pt}}%
\newcommand{\midtopline}{ %
        \arrayrulecolor{rulecolor}\specialrule{\lightrulewidth}{0pt}{0pt}}%
\newcommand{\bottomline}{ %
        \arrayrulecolor{rulecolor} \specialrule{\lightrulewidth}{0pt}{0pt}}%

\usepackage[pagebackref=true,breaklinks=true,letterpaper=true,colorlinks,bookmarks=false]{hyperref}

\definecolor{mycol1}{RGB}{209,220,229}
\definecolor{mycol2}{RGB}{237,230,221}
\definecolor{mycol3}{RGB}{227,231,197}

\definecolor{mycol4}{RGB}{242    239    249}
\definecolor{mycol5}{RGB}{220    243    242}
\definecolor{mycol6}{RGB}{222    242    217}
\definecolor{mycol7}{RGB}{247    238    237}
\definecolor{myred}{rgb}{1, 0.92, .56}

\newcommand\blfootnote[1]{%
  \begingroup
  \renewcommand\thefootnote{}\footnote{#1}%
  \addtocounter{footnote}{-1}%
  \endgroup
}

\hypersetup{
    urlcolor=blue,
}

\title{Flow-based Generative Models for Learning Manifold to Manifold Mappings}
\author{
    Xingjian Zhen\textsuperscript{\rm 1}, \ \ Rudrasis Chakraborty\textsuperscript{\rm 2}, \ \ Liu Yang\textsuperscript{\rm 1}, \ \ Vikas Singh\textsuperscript{\rm 1}\\
}
\affiliations{
    \textsuperscript{\rm 1} University of Wisconsin--Madison \\ \textsuperscript{\rm 2} University of California, Berkeley  \\
    xzhen3@wisc.edu, rudrasischa@gmail.com, lyang422@wisc.edu, vsingh@biostat.wisc.edu
  \blfootnote{Code and supplementary materials are available at\\ \url{https://github.com/zhenxingjian/Dual_Manifold_GLOW}\\ An introduction video of this paper is available at\\ \url{https://youtu.be/0r96U0vXsCM}}
  }

\begin{document}

\maketitle
\begin{abstract}
  Many measurements or observations in computer vision and machine learning manifest as non-Euclidean data. While recent proposals (like spherical CNN) have extended a number of deep neural network architectures to manifold-valued data, and this has often provided strong improvements in performance, the literature on generative models for manifold data is quite sparse. Partly due to this gap, there are also no modality transfer/translation models for manifold-valued data whereas numerous such methods based on generative models are available for natural images. This paper addresses this gap, motivated by a need in brain imaging -- in doing so, we expand the operating range of certain generative models (as well as generative models for modality transfer) from natural images to images with manifold-valued measurements. Our main result is the design of a two-stream version of GLOW (flow-based invertible generative models) that can synthesize information of a field of one type of manifold-valued measurements given another. On the theoretical side, we introduce three kinds of invertible layers for manifold-valued data, which are not only analogous to their functionality in flow-based generative models (e.g., GLOW) but also preserve the key benefits (determinants of the Jacobian are easy to calculate). For experiments, on a large dataset from the Human Connectome Project (HCP), we show promising results where we can reliably and accurately reconstruct brain images of a field of orientation distribution functions (ODF) from diffusion tensor images (DTI), where the latter has a $5\times$ faster acquisition time but at the expense of worse angular resolution.  
\end{abstract}

\section{Introduction}\label{intro}
Many measurements in computer vision and machine learning appear in a form that does not satisfy
common Euclidean geometry assumptions. Operating on data where the data samples live in structured spaces 
often leads to situations where even simple operations such as distances, angles and inner products
need to be redefined: while occasionally,
Euclidean operations may suffice, the error progressively increases 
depending on the curvature of the space at hand \cite{feragen2015geodesic}. 
One encounters such data quite often --
shapes \cite{chang2015shapenet}, surface normal directions \cite{straub2015dirichlet}, graphs and trees \cite{scarselli2008graph,kipf2016semi}
as well as probability distribution functions \cite{srivastava2007riemannian}
are some common examples in
vision and computer graphics \cite{bruno2005mesh,huang2019operatornet}. Symmetric positive definite matrices \cite{moakher2005differential,jayasumana2013kernel}, rotation matrices \cite{kendall2017geometric}, samples from a sphere \cite{koppers2016direct}, subspaces/ Grassmannians \cite{huang2018building,chakraborty2017intrinsic}, and a number of
other algebraic objects are key ingredients
in the design of efficient algorithms in computer vision and medical image analysis
as well as in the development or theoretical analysis of various machine learning problems.
While a mature literature on extending classical models such as principal components analysis \cite{dunteman1989principal},
Kalman filtering \cite{haykin2004kalman,grewal2011kalman}, regression \cite{fletcher2013geodesic} to such a manifold data regime is available,
identifying ways in which deep neural network (DNN) models can be adapted to leverage and utilize the geometry
of such data has only become a prominent research topic recently
\cite{bronstein2017geometric,chakraborty2018manifoldnet,kondor2018generalization,huang2018building,huang2017riemannian}.
This research direction has already provided convolutional
neural networks for various types of manifold measurements \cite{masci2015geodesic,masci2015shapenet} as well as sequential models such as LSTM \cite{hochreiter1997long}/GRU \cite{cho2014properties} for manifold settings \cite{jain2016structural,pratiher2018grading,chakraborty2018statistical,zhen2019dilated}.

The results in the literature, so far, on harnessing the power
of DNNs for better analysis of manifold or structured data are impressive,
but most approaches are discriminative in nature. In other words, the goal is to characterize
the conditional distribution $P(Y|\phi(X))$ based on the predictor variables or features $X$, here $X$ is
manifold-valued and the responses or labels $Y$ are Euclidean. The technical thrust is on
the design of mechanisms to specify $\phi(\cdot)$ so that it respects the geometry of the
data space.
In contrast, work on the generative side is very sparse, and to our knowledge, only a couple of methods 
for a few specific manifolds have been proposed thus far \cite{brehmer2020flows,rey2019diffusion,miolane2020learning, huang2019manifold}. As a result, the numerous application settings where generative models have shown
tremendous promise, namely, semi-supervised learning, data augmentation \cite{antoniou2017data,radford2015unsupervised}
and synthesis of new image
samples by modifying a latent variable \cite{kingma2018glow,sun2019dual} as well as numerous others, currently cannot be easily evaluated for domains with data-types that are not Euclidean vector-valued data. 

\noindent
{\bf GANs for Manifold data: what is challenging?}
There are some reasons why generative models have sparingly been applied to manifold data.
A practical consideration is that many application areas where manifold data are common, such as
shape analysis and medical imaging, cannot often provide the sample sizes needed to train
off-the-shelf generative models such as Generative adversarial networks (GANs) \cite{goodfellow2014generative} and Variational auto-encoders (VAEs) \cite{kingma2013auto,doersch2016tutorial}. There are also several issues on the technical side.
Consider the case where a data sample corresponds to an image where each pixel is a manifold variable (such as
a covariance matrix). This means that each sample lives on a product space of the manifold of covariance
matrices. 
In attempting to leverage state of the art methods for
GANs such as Wasserstein GANs (WGANs) \cite{arjovsky2017wasserstein} will involve, as a first step, 
defining appropriate generators that take uniformly distributed samples on a product space of manifolds
and transforming it into ``realistic'' samples which are also samples on a product space of manifolds.
In principle, this can be attempted via recent developments by extending spherical CNNs or other
architectures for manifold data \cite{chakraborty2018cnn}. Next, one would not only
need to define optimal transport \cite{fathi2010optimal} or Wasserstein distances \cite{huang2019manifold} in complicated spaces, but also develop
new algorithms to approximate such distances (e.g., Sinkhorn iterations) to make the overall procedure
computationally feasible.
An interesting attempt to do so was described in \cite{huang2019manifold}. In that paper, Huang et al. introduced a WGAN-based generative model that can generate low-resolution low-dimension manifold-valued images.
On the other hand, VAEs are mathematically more convenient in comparison for
such data, and as a result, a few recent works show how they can be used for dealing with
manifold-valued data \cite{miolane2020learning}. While these methods inherit VAE's advantages such as ease of synthesis,
VAEs are known to suffer from optimization challenges as well as a tendency to generate smoothed
samples. It is not clear how the numerical issues, in particular,
will be amplified once we move to manifold data where the core operations of calculating
geodesics and distances, evaluating derivatives, and so on, must also invoke numerical optimization routines.

\vspace*{0.1cm}
\noindent
{\bf Contributions. } Instead of GANs or VAEs, the use of flow-based generative models \cite{rezende2015variational,kingma2018glow}, will enable
latent variable inference and log-likelihood evaluation. It turns out, as we will show in our
development shortly, that the key components (and layers) needed in flow-based generative models with
certain mathematical/procedural adjustments, extends nicely to the manifold setting.
The goal of this work is to describe our theoretical developments and show promising experiments
in brain imaging applications involving manifold-valued data.

\section{Preliminaries}
\label{sec:prelim} 
This subsection briefly summarizes some differential geometric concepts/notations we will use.
The reader will find a more comprehensive treatment in \cite{boothby1986introduction}. 

\begin{definition}[Riemannian manifold and metric]
  Let $(\mathcal{M},g^\mathcal{M})$ be an orientable complete Riemannian manifold with a Riemannian metric $g$, i.e., $\forall x \in \mathcal{M}: g_x:T_x{\mathcal{M}}\times T_x{\mathcal{M}} \rightarrow \mathbf{R}$ is a bi-linear symmetric positive definite map, where $T_x\mathcal{M}$ is the tangent space of $\mathcal{M}$ at $x\in \mathcal{M}$. Let $d: \mathcal{M} \times \mathcal{M} \rightarrow [0,\infty)$ be the distance induced from the Riemannian metric $g$.
\end{definition}

\begin{definition}
\label{theory:def1}
Let $p \in \mathcal{M}$, $r > 0$. Define $\mathcal{B}_r(p) =
\left\{ q \in \mathcal{M} | d(p,q) < r \right\}$ to be an open ball at
$p$ of radius $r$.
\end{definition}

\begin{definition}[Local injectivity radius \cite{groisser2004newton}]
\label{theory:def2}
The local injectivity radius is defined as $r_{\text{inj}}(p) = \sup \left\{ r
| \text{Exp}_p : (\mathcal{B}_r(\mathbf{0}) \subset T_p \mathcal{M} )
\rightarrow \mathcal{M} \right\}$ where $\text{Exp}_p$ is defined and is a
diffeomorphism onto its image at $p \in \mathcal{M}$.
The {\it injectivity radius} \cite{manton2004globally} of
$\mathcal{M}$ is defined as $r_{\text{inj}}(\mathcal{M}) =
\inf_{p \in \mathcal{M}} \left\{r_{\text{inj}}(p)\right\}$.
\end{definition}

\begin{figure}[!b]
        \centering
               \includegraphics[width=0.75\columnwidth]{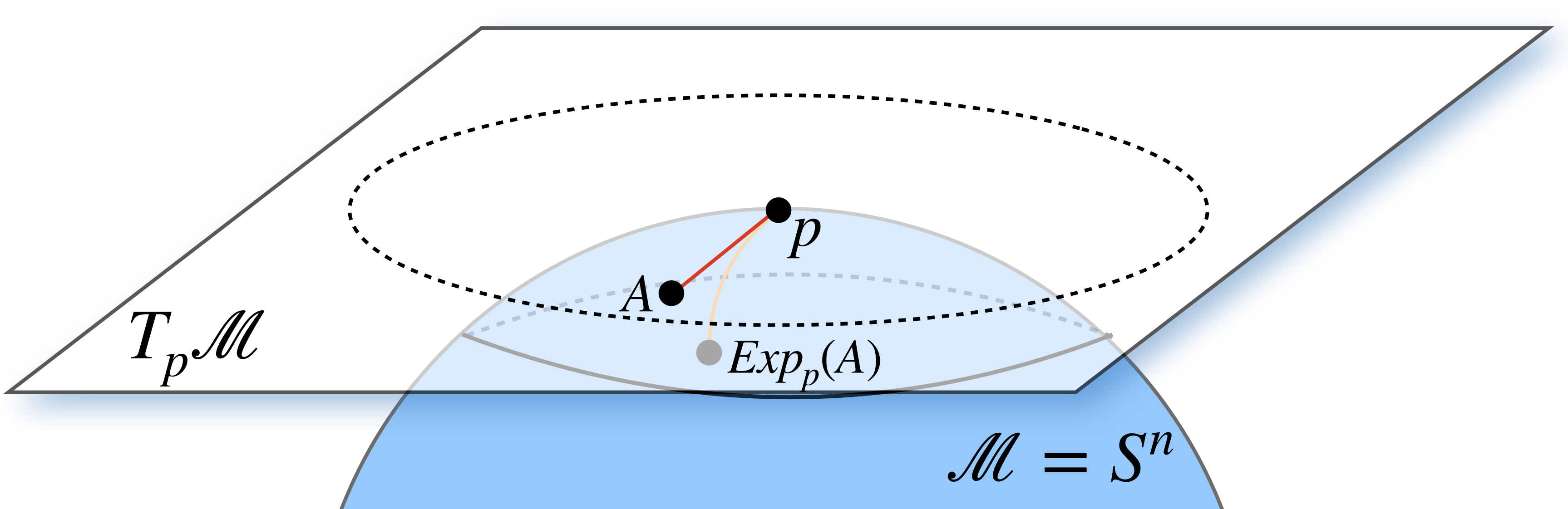}
               \caption{Schematic description of an exemplar manifold ($\mathbf{S}^n$) and the corresponding tangent space at a ``pole''. }\label{tangent}
\end{figure}

Within $\mathcal{B}_r(p)$, where $r \leq r_{\text{inj}}(\mathcal{M})$,
the mapping $ \text{Exp}^{-1}_p: \mathcal{B}_r(p) \rightarrow
\mathcal{U} \subset T_p \mathcal{M} \subset \mathbf{R}^m$, is called the inverse
Exponential/Log map, $m$ is the dimension of $\mathcal{M}$. For each point $p \in \mathcal{M}$, there exists an open ball $\mathcal{U} =  \mathcal{B}_r(q)$ for some $q\in \mathcal{M}$ such that $p \in \mathcal{U}$, where $r = r_{\text{inj}}(\mathcal{M})$. Thus, we can cover $\mathcal{M}$ by an indexed (possibly infinite) cover $\left\{\left(\mathcal{B}_r(q), \text{Exp}^{-1}_q\right)\right\}_{q\in \mathcal{I}}$. This set is an example of a {\it chart} on $\mathcal{M}$; for an example, see \cite{krauskopf2007numerical} and also Fig. \ref{tangent}.

For notational simplicity, we will denote a chart covering $p \in \mathcal{M}$ by $\Phi_p$, since in general, we can use an arbitrary chart instead of an inverse Exponential map. Note that the domain for two chart maps may not necessarily be disjoint.  

Given a differentiable function $F:\mathcal{M} \rightarrow \mathcal{M}$ defined as $x \mapsto \Psi^{-1}\left(\widetilde{F}\left(\Phi(x)\right)\right)$, where $\Phi$ and $\Psi$ are the functions in the chart covering $x$ and $F(x)$ respectively and for some differentiable $\widetilde{F}: \mathbf{R}^m \rightarrow \mathbf{R}^m$, the Jacobian of $F$ (denoted by $\tenq{\frac{dy}{dx}}$) is defined as:
{
\begin{align}
\tenq{\frac{dy}{dx}} := \frac{\partial \Psi\circ \Phi^{-1}}{\partial \Phi(x)} \frac{d\widetilde{F}}{d\widetilde{x}}\Big\vert_{\Phi(x)}
\label{chart_d}
\end{align}}

The reason for the peculiar notation is that the derivative cannot be defined on manifold-valued data, so $\frac{dy}{dx}$ is not meaningful: we use the notation $\tenq{\cdot}$ to acknowledge this difference. 
Also note that $\Psi,\Phi$ are the same only when \begin{inparaenum}[\bfseries (1)] \item using the global charts for space $X$ and $F(X)$ \item $X$ and $F(X)$ are on the same manifold.\end{inparaenum} 
\begin{definition}[{\bf Group of isometries of $\mathcal{M}$} ($I\left(\mathcal{M}\right)$)]
\label{theory:def6}
{A diffeomorphism $\iota: \mathcal{M} \rightarrow \mathcal{M}$ is an
  isometry if it preserves distance, i.e., $d\left(\iota\left(x\right),
  \iota\left(y\right)\right) = d\left(x, y\right)$. The set
  $I(\mathcal{M})$ 
  forms a group
  with respect to function composition.}
\end{definition}

Rather than writing an isometry
  as a function $\iota$, we will write it as a group action.
  Henceforth, let $G$ denote the group $I(\mathcal{M})$, and for $g
  \in G$, $x \in \mathcal{M}$, let $g\cdot x$ denote the result of
  applying the isometry $g$ to point $x$. Similar to the terminologies in \cite{chakraborty2018statistical}, we will use the term ``translation'' to denote the group action $\iota$. This is due to the distance preserving property and is inspired by the analogy from the Euclidean space.

\section{Flow-based Generative Models}
\label{sec:flow-model} 

In this section, we will introduce flow-based generative models for manifold-valued data. We will first describe the Euclidean formulation and specify which components need to be generalized to get the manifold-valued formulation. 

\subsection{Flow-based Models: Euclidean Case}
Flow-based generative models \cite{rezende2015variational,kingma2018glow,yang2019pointflow} aim to maximize the log-likelihood of the input data from an unknown distribution.
The idea involves mapping the {\em unknown distribution in the input space} to a {\em known distribution in the latent space} using an invertible function, $f$. At a high level, sampling from a known distribution is simpler, so an 
invertible $f$ can help draw samples from the input space distribution. 

Let $\left\{\mathbf{x}_i\right\}$ be i.i.d. samples drawn from an unknown distribution $p^*\left(\mathbf{x}\right)$. Let this unknown distribution be parameterized by $\theta$. In the rest of the paper, we use $p_{\theta}(\mathbf{x})$ as a proxy for  $p^*\left(\mathbf{x}\right)$. We learn $\theta$ over a dataset $\mathcal{D}$. We maximize the likelihood of the model $\theta$ given the dataset $\mathcal{D}$ by minimizing the equivalent formulation of negative log-likelihood as:

\begin{align}
    {\mathcal \ell(\theta|\mathcal D)} &= \frac{1}{N}\sum_{i=1}^{N}-{\rm log}p_\theta ({\mathbf x}_i)
\end{align}

But to minimize the above expression, we need to know $p_{\theta}$. One way to bypass this problem is to learn a mapping from a {\rm known} distribution in the latent space. Let the latent space be $\mathbf{z}$. Then, the generative step is given by ${\mathbf z} \sim p({\mathbf z}),\ {\mathbf x} = {g}({\mathbf z})$. 
Here $p(\mathbf{z})$ can be a Gaussian distribution $\mathcal{N}\left(\mathbf{z};\mathbf{0},I\right)$.

Let $f$ be the inverse of $g$.
For normalizing flow \cite{rezende2015variational}, $f$ is composed as a sequence of invertible functions ${f}=f_1\circ f_2 \circ \ldots \circ f_K$. Hence, we have

\begin{align*}
    \mathbf x \stackrel{f_1}{\longleftrightarrow} \mathbf h_1 \stackrel{f_2}{\longleftrightarrow} \mathbf h_2 \ldots \stackrel{f_K}{\longleftrightarrow} \mathbf z
\end{align*}
Using $\mathbf h_0=\mathbf x$ and $\mathbf h_{K}=\mathbf z$, the log-likelihood of $p_\theta (\mathbf x)$ is
{
\begin{align}
\label{eq:glow}
    {\rm log}p_\theta (\mathbf x) &= {\rm log}p(\mathbf z) + {\rm log}|{\rm det}(d\mathbf z / d\mathbf x)| \\
                                  &= {\rm log}p_\theta (\mathbf z) + \sum_{j=0}^{K}{\rm log}|{\rm det}(d\mathbf h_{j} / d\mathbf h_{j-1})| \numberthis
\end{align}}
In \cite{kingma2018glow}, the GLOW model is composed of three different layers whose Jacobian $d\mathbf h_{j} / d\mathbf h_{j-1}$ is a triangular matrix, simplifying the log-determinant:
{
\begin{align}
    {\rm log}|{\rm det}(d\mathbf h_{j} / d\mathbf h_{j-1})| = {\rm sum}({\rm log}|{\rm diag}(d\mathbf h_{j} / d\mathbf h_{j-1})|)
\end{align}}

\begin{figure}[!b]
\centering
\includegraphics[width=0.95\columnwidth]{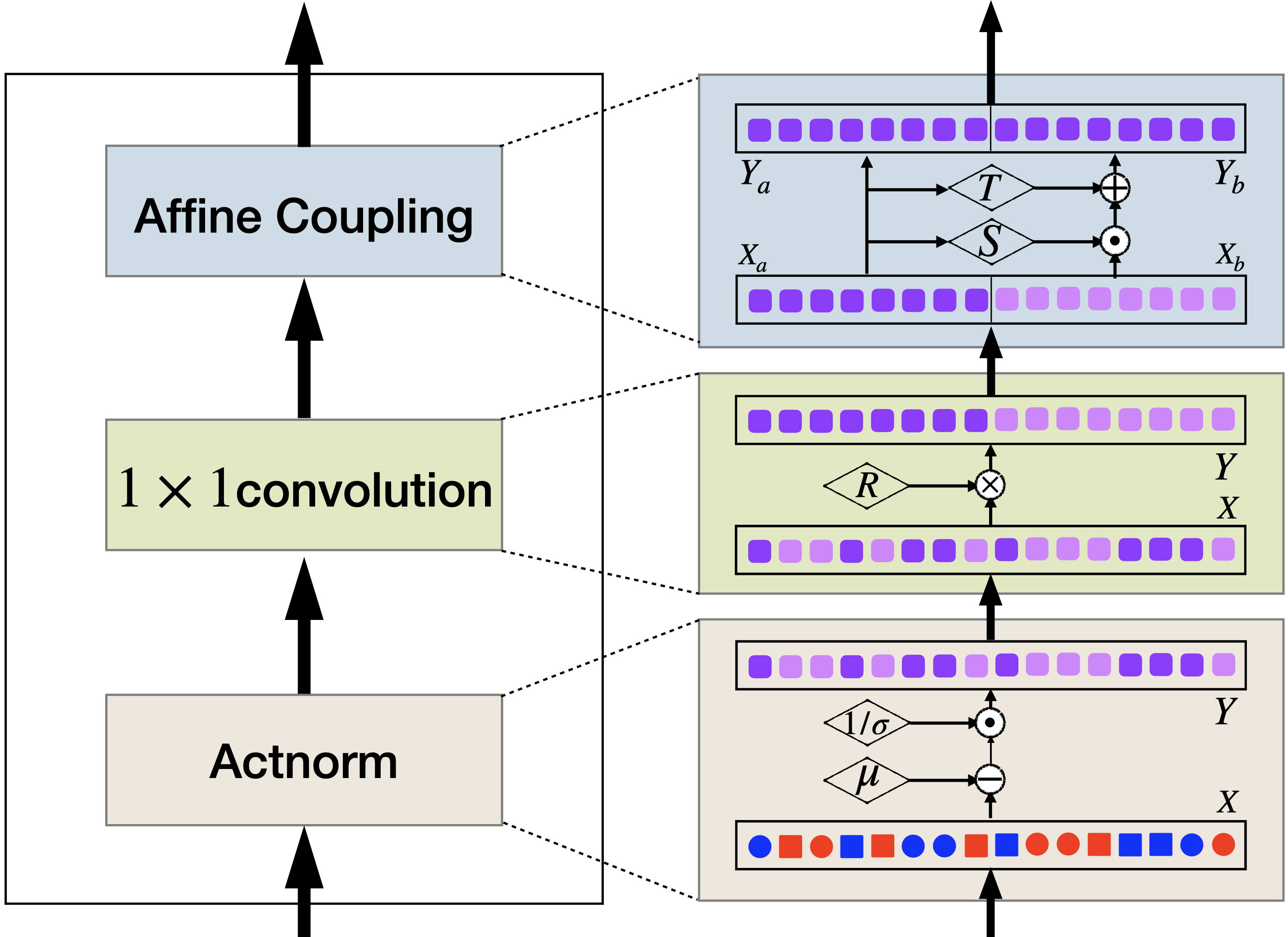}
\caption{\footnotesize The basic block of GLOW \cite{kingma2018glow}. The color represents the mean while the shape represents the standard deviation. The target distribution on the latent space is the ``\textcolor[RGB]{134,73,245}{Grape}'' rounded rectangles. {\it Actnorm} normalizes the data to almost ``\textcolor[RGB]{134,73,245}{Grape}'' rounded rectangles,
while the disturbance part is ``\textcolor[RGB]{202,139,248}{Lavender}''. {\it$1 \times 1$ convolution} organizes the channels. {\it Affine Coupling} operates on half of the channels to fit the target distribution. The ``$\cdot$'' here is the element-wise multiplication, while ``$\times $'' is the matrix multiplication. ``$+/-$'' are of the normal definition.}\label{block}
\end{figure}
The three layers in the basic GLOW block (shown in Fig. \ref{block}), summarized in Table \ref{glowFunction} are all invertible functions. These are (a) {\it Actnorm}, (b) {\it Invertible $1\times 1$ convolution}, and (c) {\it Affine Coupling} Layers. 
Note that the data is squeezed before it is fed into the block.
Then, the data is split as in \cite{dinh2016density}.

\noindent
{\bf (a) Actnorm} normalizes the input to be a zero-mean and identity standard deviation.
In \eqref{eq:actnorm_glow}, $\mu,\sigma$ are initialized from the data and then trained independently.

\noindent
{\bf (b) $1\times 1$ convolution} applies the invertible matrix $R$ on the channel dimension. In \eqref{eq:1x1conv_glow}, $X\in \mathbf{R}^{s_r\times c_r}$ and $R\in \mathbf{R}^{c_r\times c_r}$ where $s_r$ is the resolution of the input variables while $c_r$ is the number of channels.

\noindent
{\bf (c) Affine Coupling} uses the idea of 
split+concatenation. In \eqref{eq:affine_glow}, the input variable $X$ is {\em split} along the channel to $X_a,X_b$, and then $Y_a,Y_b$ are 
{\em concatenated} to get the final output $Y$. Here, $S$ (and $T$) are real-valued matrices of the same dimension as $X_b$ for element-wise scaling (and translation).

\begin{figure*}[!t]
        \centering
                \includegraphics[width=1.0\textwidth]{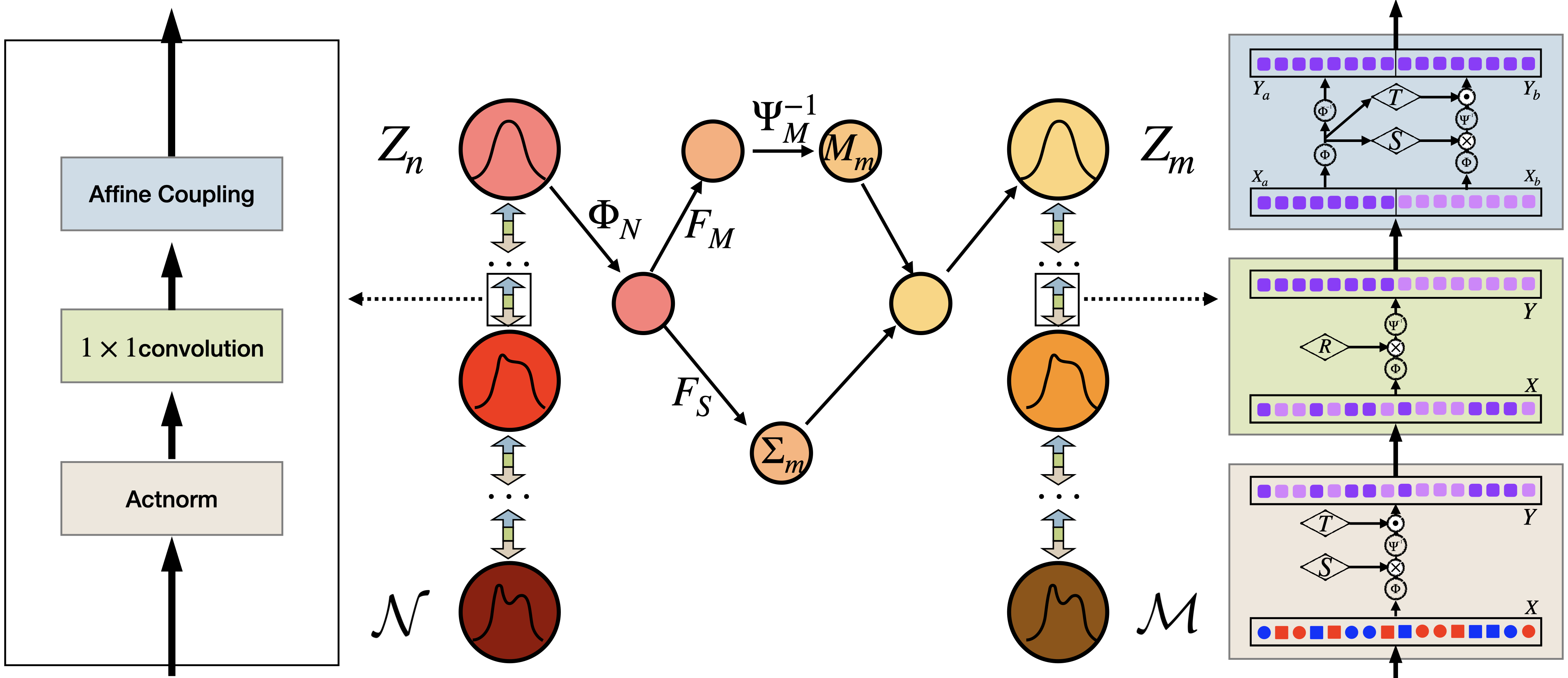}
               \caption{Transfer from the source manifold $\mathcal{N}$ to the target manifold $\mathcal{M}$ with the generative model. The detail of blocks of our model. The meanings of colors and shapes are the same as Fig. \ref{block}, while all variables lie on the manifold instead of Euclidean space. The major difference between our manifold-valued GLOW and the original GLOW model is we use a tangent space transformation before and after every operator. Different from Fig. \ref{block}, there is no element-wise multiplication. The ``$\cdot$'' here is the group operation on the manifold-valued data. The ``$\times $'' is also the matrix multiplication in the tangent space. }\label{whole_structure}
\end{figure*}

\begin{table}[!t]
  \begin{center}
    \scalebox{0.74}{
    \begin{tabular}{ >{\columncolor{mycol2}}c|>{\columncolor{mycol3}}c|>{\columncolor{mycol1}}c}
      \topline
       Actnorm & $1 \times 1$ convolution & Affine Coupling \\
      \midtopline
			\parbox{3.2cm}{\begin{equation}\label{eq:actnorm_glow}Y = \frac{1}{\sigma}\odot(X-\mu)\end{equation}}
		&
			\parbox{2.8cm}{\begin{equation}\label{eq:1x1conv_glow}Y = R\times X\end{equation}}
		&
			\parbox{4cm}{\begin{equation}\label{eq:affine_glow} \begin{split} S, T &= \textsf{NN}\left(X_a\right) \\ Y_b &= S\odot X_b+T \\ Y_a &= X_a \end{split}\end{equation}}\\
	 \bottomline
    \end{tabular}}
  \end{center}
\caption{\label{glowFunction}Definition of {\it Actnorm}, {\it $1 \times 1$ convolution} and {\it Affine Coupling} layers in basic GLOW block. $\odot$ is the elementwise multiplication.The function \textsf{NN}() is a nonlinear mapping.}

\end{table}

In \cite{kingma2018glow}, authors use a closed form for the inverse of these layers. 
Notice that calculating the determinant of the Jacobian is simple for all these layers except the affine coupling layer in \eqref{eq:affine_glow} (Table \ref{glowFunction}). 

Since $\frac{dY_a}{dX_b}=0,\frac{dY_a}{dX_a}=I$, the Jacobian determinant is {$\text{det}(S)$}.

\begin{align*}
\label{eq:detjac_glow}
     \text{det}\left(\frac{dY}{dX}\right)&= \text{det}\left( \begin{bmatrix}
                       \frac{dY_a}{dX_a}&\frac{dY_a}{dX_b}\\
                       \frac{dY_b}{dX_a}&\frac{dY_b}{dX_b}
                   \end{bmatrix} \right)  \\ &= 
                   \text{det}\left( \begin{bmatrix}
                       I & 0 \\
                       \frac{dY_b}{dX_a}& S 
                   \end{bmatrix} \right) = \text{det}\left( S \right) \numberthis 
\end{align*}

\noindent
{\bf Next Steps:} With the description above, we can now list the key operational components in \eqref{eq:actnorm_glow}-\eqref{eq:detjac_glow}, which we need to modify for our manifold-valued extension.

\vspace*{0.1cm}
\noindent
{\bf Key ingredients:} 
In \eqref{eq:actnorm_glow} and \eqref{eq:affine_glow},
the operators are 
\begin{inparaenum}[\bfseries (i)] 
\item elementwise multiplication for $\sigma, S$ and 
\item the addition of bias for $\mu,T$. 
\item In
\eqref{eq:1x1conv_glow}, we require invertible matrices. 
\item Finally, to compute the log-likelihood, we need the calculation of derivative in \eqref{eq:detjac_glow}.
\end{inparaenum}
Thus we can verify that the key ingredients to define the model in GLOW are
\begin{inparaenum}[\bfseries (i)] \item elementwise multiplication; \item
    addition of bias; \item invertible matrix; \item derivative calculation. 
\end{inparaenum} 
In theory, if we can modify those components from Euclidean space to manifolds, we will obtain a flow-based generative model on a Riemannian manifold. 
Observe that {\bf (i)} and {\bf (iii)} are matrix multiplications, which are non-trivial to define on a manifold. 
In Def. \ref{theory:def2}, we can use the chart map to map the manifold to a subspace of $\mathbf R^m$ where a matrix multiplication can be used.
This also provides a way to solve item {\bf (iv)} based on the chart map. In \eqref{chart_d}, we show how to compute the Jacobian of a differentiable function $F$ from one manifold to another, respecting to the charts of the manifolds.
For the item {\bf (ii)}, adding a bias can be viewed as a ``translation'' in the Euclidean space, while in Def. \ref{theory:def6} we define the translation on manifold-valued data using the group action. With these in hand, we are ready to present our proposed manifold version of these layers next. 

\subsection{Flow-based Models: Riemannian Manifold Case}
\label{glowman}

\begin{table*}[t]
  \begin{center}
   \scalebox{0.9}{
    \begin{tabular}{>{\columncolor{mycol2}}c|>{\columncolor{mycol3}}c|>{\columncolor{mycol1}}c}
      \topline
       Actnorm & $1 \times 1$ convolution & Affine Coupling \\
      \midtopline
			\parbox{5.7cm}{\begin{equation}\label{eq:act_man}Y =  \Psi^{-1}\left(S\times \Phi\left(X\right)\right) \cdot T \end{equation}}
		&
			\parbox{4.8cm}{\begin{equation}\label{eq:conv_man}Y =  \Psi^{-1}\left(R\times \Phi\left(X\right)\right)\end{equation}}
		&
			\parbox{5.9cm}{{\small \begin{equation}\label{eq:couple_man}\begin{split} S, T &= \textsf{NN}\left(\Phi(X_a)\right) \\ Y_b &= \Psi^{-1}\left(S\times \Phi\left(X_b\right)\right) \cdot T \\ Y_a &= X_a \end{split}\end{equation}}}\\
     \midtopline
			\parbox{5.7cm}{\begin{equation}\label{eq:invact_man}X = \Phi^{-1}\left(S^{-1} \times  \Psi\left(Y \cdot T^{-1}\right)\right)\end{equation}}
		&
			\parbox{4.8cm}{\begin{equation}\label{eq:invxonv_man}X =  \Phi^{-1}\left(R^{-1} \times \Psi\left(Y\right)\right)\end{equation}}
		&
			\parbox{5.9cm}{\begin{small} \begin{equation}\label{eq:invcouple_man}\begin{split} S, T &= \textsf{NN}\left(\Phi(Y_a)\right) \\ X_b &= \Phi^{-1}\left(S^{-1}\times \Psi\left(Y_b \cdot T^{-1}) \right)\right) \\ X_a &= Y_a \end{split}\end{equation}\end{small}}\\
	 \bottomline
    \end{tabular}
}
  \end{center}
\caption{Definition of {\it Actnorm}, {\it $1 \times 1$ convolution} and {\it Affine Coupling} layers in our ManifoldGLOW block, with forward function on the top and reverse function in the bottom. Here $\Phi$ and $\Psi^{-1}$ are the Chart Map and its inverse. $S$ is a diagonal matrix, so $S^{-1}$ can be computed elementwise. $T^{-1}$ represents the inverse of the group action. The $R$ is chosen as the rotation matrix. Thus, $R^{-1}=R^T$.}
\label{manglowFunction}
\end{table*}

We will now introduce the manifold counterpart of the key operations. See Table \ref{manglowFunction} for a
summary of functions.

\vspace*{0.1cm}
\noindent
{\bf (a) Actnorm.} Let $s_r$ be the spatial resolution and $c_r$ be the channel size, $X \in \mathcal{M}^{s_r\times c_r}$.
We modify (\ref{eq:actnorm_glow}) to manifold-valued data using the operators we mentioned above in Key ingredients. The bias term is replaced by the group operators $T$ while the multiplication $1/\sigma$ is replaced by the diagonal matrix $S$ of size $m\times m$ in the space after chart mapping $\Phi(\cdot)$. The layer function is defined as in (\ref{eq:act_man}).

\noindent
{\it Determinant of the Jacobian}
can be computed as shown below in \eqref{eq:detact_man}. In general, $s_r$ can be a tuple, i.e., for 3D data, it is a 3 dimensional tuple. 

\begin{align*}
\label{eq:detact_man}
\text{det}\left(\tenq{\frac{dY}{dX}}\right) = \prod_{s_r\times c_r} \left(\prod s_i\right) \text{det}\left(\Psi\circ \Phi^{-1}\right) \numberthis 
\end{align*}

\vspace*{0.1cm}
\noindent
{\bf (b) $1\times 1$ convolution.}
We define a $1\times 1$ convolution to offer the flexibility of interaction between channels. Here $R$ is a $c_r\times c_r$ matrix applied after chart mapping $\Phi(\cdot)$. In general, we can learn any $R\in \textsf{GL}(c_r)$, i.e., a full rank matrix like in \eqref{eq:1x1conv_glow}. But in practice, maintaining full rank is a hard constraint and may become unbounded. As a regularization, we choose $R$ to be a rotation matrix. This layer function is defined as in \eqref{eq:conv_man} using the same notation as in
\eqref{eq:1x1conv_glow}.

\noindent
{\it Determinant of the Jacobian} can be computed as shown below in \eqref{eq:detconv_man}. Notice that for $R$ to be a rotation matrix, the contribution from $\text{det}(R)$ is $\pm 1$. 
{
 \begin{align*}
\label{eq:detconv_man}
\text{det}\left(\tenq{ \frac{dY}{dX}}\right) = \prod_{s_r} \text{det}(R) \text{det}\left(\Psi\circ \Phi^{-1}\right) \numberthis
\end{align*}}

\vspace*{0.1cm}
\noindent
{\bf (c) Affine Coupling.} For manifold-valued data, 
given $X \in \mathcal{M}^{s_r\times c_r}$ (where $s_r$ and $c_r$ are spatial and channel resolutions), we first split the data along the channel dimension, i.e., partition $c_r$ into two parts denoted by $X_a \in \mathcal{M}^{s_r\times c_a}$ and $X_b \in \mathcal{M}^{s_r\times c_b}$, where $c_r = c_a + c_b$. {From \eqref{eq:affine_glow}, we need to modify the scaling and translation}. Here, $S \in \left(\mathbf{R}^{m\times m}\right)^{s_r\times c_a}$ and $T \in G^{s_r\times c_a}$. These two operators play the same roles as in \eqref{eq:affine_glow}, scaling and translation. We need $S$ to be full rank. If needed, one may use constraints like orthogonality or bounded matrix for numerical stability. After performing the coupling, we simply combine $Y_a$ and $Y_b$ to get $Y  \in \mathcal{M}^{s_r\times c_r}$ as our output. This function is defined in \eqref{eq:couple_man}. 

\noindent
{\it Determinant of the Jacobian} 
 can be computed as:
\begin{align*}
	\begin{bmatrix}
 \tenq{\frac{dY_a}{dX_a}} & \tenq{\frac{dY_a}{dX_b}} \\
 \tenq{\frac{dY_b}{dX_a}} & \tenq{\frac{dY_b}{dX_b}} 
 \end{bmatrix} \numberthis
\end{align*}
 
  Similar to \eqref{eq:detjac_glow}, observe that $ \tenq{\frac{dY_b}{dX_a}}$ involves taking the gradient of a neural network! But fortunately,
  we only require the determinant of the Jacobian matrix, and the independence of $Y_a$ on $X_b$ saves the calculation of $ \tenq{\frac{dY_b}{dX_a}}$ since $\tenq{\frac{dY_a}{dX_b}} = 0$.
Thus, given $X,Y \in \mathcal{M}^{s_r\times c_r}$, the Jacobian determinant is given as 
{
 \begin{align*}
\label{eq:detcouple_man}
\text{det}\left(\tenq{\frac{dY}{dX}}\right) = \prod_{s_r\times c_r} \text{det}\left(S\right)\text{det}\left(\Psi\circ \Phi^{-1}\right) \numberthis
\end{align*}}

\vspace*{0.1cm}
\noindent
{\bf Distribution on the latent space:}
After the cascaded functional transformations described above, we transform $X$ to the latent space $Z \in \mathcal{M}^{s_h\times c_h}$. We define a Gaussian distribution on $Z$, namely 
$P\left(Z;M,\Sigma\right)$, by inducing a
multi-variate Gaussian distribution from $\mathbf{R}^m$ as 
{
\begin{align*}
\label{eq:gauss}
 \left.\exp\left(-\frac{\left(\Phi(Z) - \Phi(M)\right)^T\Sigma^{-1}\left(\Phi(Z) - \Phi(M)\right)}{2}\right) 
 \middle/ C(\Sigma) \right.\numberthis
\end{align*}}
where$M \in \mathcal{M}^{s_h\times c_h}$ and $\Sigma \in \textsf{SPD}(m)^{s_h\times c_h}$ ($\textsf{SPD}$ denotes a symmetric positive definite matrix). $C(\Sigma)$ is the normalization constant to make the total probability to be $1$.

\subsection{Learning Mappings Between Manifolds}

We can now ask the question: {\it can we draw manifold-valued data conditioned on another manifold-valued sample?} Due to the nature of the invertibility of our generative model, this seems to be possible since all we need to develop, in addition to what has been covered, is a scheme to sample
data from Euclidean space conditioned on a vector-valued input. 

Recently, extensions of the GLOW model (in a Euclidean setup) have been used to generate samples from space $\mathcal{X}$ conditioned on space
$\mathcal{Y}$, see \cite{sun2019dual}. In this section, we roughly follow \cite{sun2019dual} 
by using connections in a latent space but in a manifold setting to generate a sample from a manifold $\mathcal{M}$, conditioned on a sample on manifold $\mathcal{N}$.
The underlying assumption is that there exists a (smooth) function from $\mathcal{N}$ to $\mathcal{M}$. The generation steps are as follows. 

\begin{asparaenum}[\bfseries Step (a):]
\item Given variables $X \in \mathcal{M}^{s_x\times c_x}$ and $Y \in \mathcal{N}^{s_y\times c_y}$ with the dimension of the manifolds $\mathcal{M}$ and $\mathcal{N}$ to be $m$ and $n$ respectively, we use the two parallel GLOW models (as discussed above) to get the corresponding latent space. Let it be denoted by $Z_m$ and $Z_n$ respectively. 
\item After getting the respective latent spaces, we need to fit a distribution on it. Since we wish to generate samples from $\mathcal{M}$, the distribution on the respective latent space $Z_m$ must be induced from the variables in $Z_n$, i.e., the latent space for $\mathcal{N}$. We do not have any constraint on the distribution parameters for $Z_n$, so, we use a Gaussian distribution with a fixed $M$ and $\Sigma$ on $Z_n$. The parameters for the Gaussian distribution on $Z_m$ are defined as functions of $Z_n$. Formally, we define $P\left(Z_m;M_m,\Sigma_m\right)$ using \eqref{eq:gauss}, where, $M_m = \Psi^{-1}_M\left(F_M(\Phi_N(Z_n)\right)$ and $\Sigma_m = F_S(\Phi_N(Z_n))$. Here, the two functions $F_M$ and $F_S$ are modeled using a neural network. The scheme is shown in Fig. \ref{whole_structure}.
\end{asparaenum}
 
\vspace*{0.1cm}
\noindent
{\bf Specific examples of manifolds.}
Finally, in order to implement \eqref{eq:act_man}, \eqref{eq:conv_man} and \eqref{eq:couple_man} mentioned in the previous sections,
 basic operations specific to a manifold are \begin{inparaenum}[\bfseries (a)] \item the choice of distance, $d$, \item the isometry group, $G$, \item the chart map $\Phi$ and its inverse, $\Phi^{-1}$.\end{inparaenum} We use three types of non-Euclidean Riemannian manifolds in the experiments presented in this work (including the supplement section), they are \begin{inparaenum}[\bfseries (a)] \item hypersphere, $\mathbf{S}^{n-1}$ \item space of positive real numbers, $\mathbf{R}_{+}$ \item space of $n\times n$ symmetric positive definite matrices ($\textsf{SPD}(n)$). \end{inparaenum} We give the explicit formulation for the operations in Table \ref{manifoldExample}.

\begin{table}[!b]
    \begin{center}
    \scalebox{0.75}{
    \begin{tabular}{ l|l|l|l}
      \topline\myrowcolour
       & $\mathbf{S}^{n-1}$ & $\mathbf{R}_{+}$ & $\textsf{SPD}(n)$ \\
       \midtopline
      $d(X,Y)$ &  $\arccos\left(X^TY\right)$ & $|\log(X/Y)|$ & $\|\textsf{logm} X^{-1}Y\|$ \\
      \midtopline
      $G$ & $\textsf{SO}(n-1)$ & $\mathbf{R}\setminus \left\{0\right\}$ & $\textsf{SO}(m)$  \\
      \midtopline
      $\Phi(X)$ & $\frac{\theta}{\sin(\theta)}\left(X - P\cos(\theta)\right)$ & $\log(X)$ & $\textsf{Chol}(X)$ \\
      \midtopline
      $\Phi^{-1}(\mathbf{v})$ & $P\cos(\|\mathbf{v}\|) + \frac{\mathbf{v}}{\|\mathbf{v}\|} \sin(\|\mathbf{v}\|)$ & $\exp(\mathbf{v})$ & $\mathbf{v}\mathbf{v}^T$\\
      \bottomline
    \end{tabular}
    }
  \end{center}
\caption{The explicit formulation for the basic operations. Here $P$ is an anchor point for chart map, which can be one of the poles. $\theta = d(P, X)$, and $\textsf{SO}(m)$ is the group of $m\times m$ special orthogonal matrices. $\mathbf{v} \in \mathbf{R}^{n-1}$. $\textsf{Chol}$ is the Cholesky decomposition.}
\label{manifoldExample}
\end{table}

\section{Experiments}\label{results}

We demonstrate the experimental results of our model using two setups. First, we generate texture images based on the local covariances, which serves as a sanity check 
evaluation relative to another  
generative model for manifold-valued data available 
at this time. The second experiment, which is our main scientific focus,  generates orientation distribution function (ODF) images  \cite{hess2006q} using diffusion tensor imaging (DTI) \cite{basser1994mr,alexander2007diffusion}. {\it Note that, in this setting we construct the DTI scans from under-sampled diffusion directions. }

\vspace*{0.1cm}
\noindent
{\bf Baseline.} 
Very recently, the $\mathcal{M}_e$-flow \cite{brehmer2020flows} was introduced, which provides a generative model for manifold-valued data. $\mathcal{M}_e$-flow uses 
an encoder to encode the manifold-valued data in the high-dimensional space into a low-dimensional Euclidean space. 
During generation, the model will generate the low-dimensional Euclidean data and warp it back to the manifold in the high-dimensional space. The benefit of this method is that it can learn the dimension of the unknown manifold, including natural images like ImageNet \cite{deng2009imagenet}. But for a known Riemannian manifold, the dimension $d$ of the manifold is fixed. For example, $\textsf{SPD}(n)$ is of dimension $d=n(n+1)/2$, while $\mathbf{S}^{m}$ is of dimension $d=m$. Thus, for a known Riemannian manifold, $\mathcal{M}_e$-flow learns the chart using an encoder neural network and applies all the operations in the learned space with (known) dimension $d$. 
Another interesting recent proposal, manifoldWGAN, 
\cite{huang2019manifold} showed that it is possible to generate $32\times 32$ resolution $\textsf{SPD}(3)$ matrices using WGAN.
Due to the involved calculations needed by WGAN, extending it into high-dimension manifold-valued data including ODF ($\mathbf{S}^{m}$) will
require non-trivial changes. 
Further, manifoldWGAN in its current form does not deal with conditioning
the generated images based on another manifold-valued data but is an interesting future direction to explore.

Now, we present experiments for generating texture images before moving to the more challenging ODF generation task. 
\subsection{Generating Texture Images}
\begin{figure}[!hbt]
        \centering
               \includegraphics[width=0.99\columnwidth]{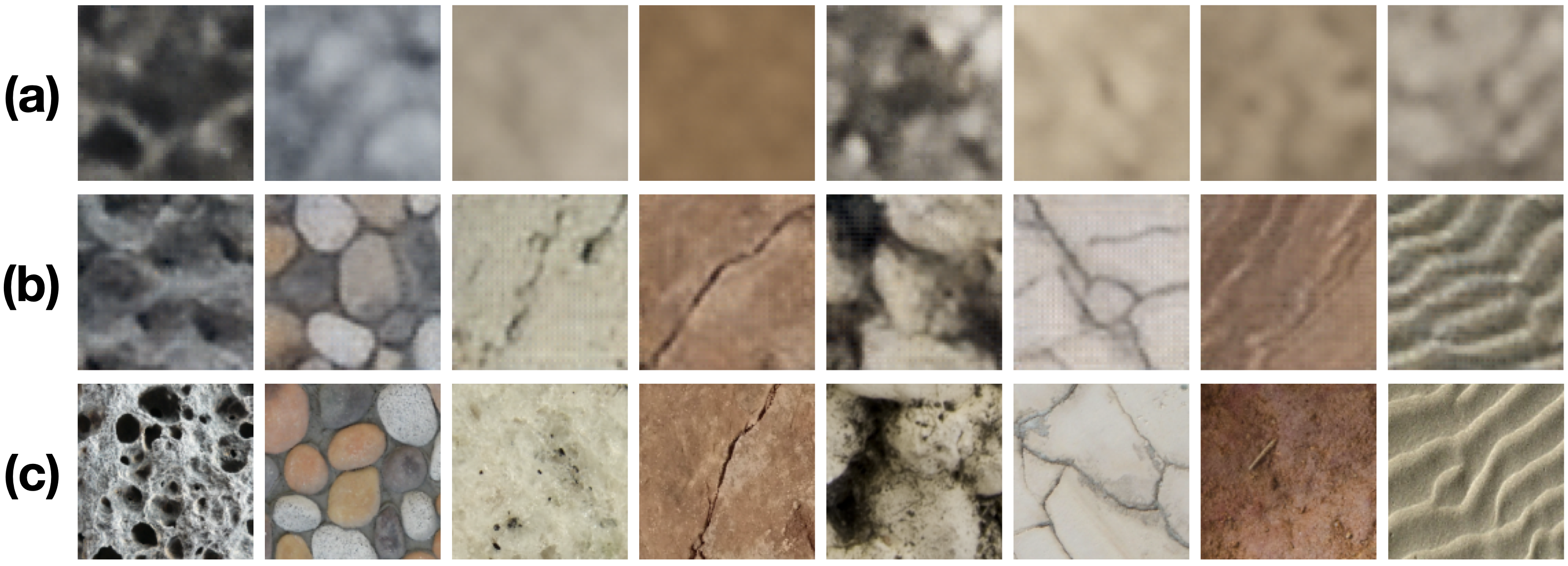}
               \caption{Generated images from (a) $\mathcal{M}_e$-flow, (b) ours, and (c) the ground truth. The condition is the local covariances of the RGB channels.}\label{texture}
\end{figure}

The earth texture images dataset was introduced in \cite{yu2019texture}. The train (and test) set have $896$ (and $98$) images. All images are augmented by random transformations and cropping to size $64\times 64$. Our goal here is to generate texture images based on the local covariances of the three (R, G, B) channels. So the two manifolds are $\textsf{SPD}(3)$ (for covariance matrix) and $\mathbf{R}_+^3$ (for texture images). Since $\mathcal{M}_e$-flow can only take the Euclidean data as the ``conditioning variable'', 
we vectorize the local covariances as the condition variable for $\mathcal{M}_e$-flow. The dimension of the learned space for $\mathcal{M}_e$-flow is chosen as $64$ (default configuration from StyleGAN \cite{karras2020analyzing}). For our case, we build two parallel manifold-GLOW with $8$ blocks on each side. After every $2$ blocks, the spatial resolution is reduced to half. In the latent space, we train a residual network with $3$ residual blocks to map the distribution of the $\textsf{SPD}(3)$ to $\mathbf{R}_+^3$. Example results are shown in Fig. \ref{texture}. 
Even in this simple setting, due to the encoder in the $\mathcal{M}_e$-flow, the generated images lose sharpness. Our model uses the information of the local covariances to generate  
superior texture images. 

\subsection{Main Focus: Diffusion MRI Dataset}
Our main focus is the conditional synthesis of structural brain image data. Diffusion-weighted magnetic resonance imaging (dMRI) is an MR imaging modality which measures the diffusion of water molecules at a voxel, and is used to understand brain structural connectivity.
Diffusion tensor imaging (DTI), a type of dMRI \cite{basser1994mr,alexander2007diffusion},
measures the restricted diffusion of water along only three canonical directions 
at each voxel.  
The measurement at each voxel is a symmetric positive definite ($\textsf{SPD}$) matrix (i.e., manifold-valued data).
If multi-shell acquisition capabilities are available, we can obtain a richer acquisition; here, each voxel is an orientation distribution function (ODF) \cite{hess2006q} which describes the diffusivity in multiple directions (less lossy compared to DTI). By symmetrically/equally sampling $362$ points on the continuous distribution function \cite{garyfallidis2014dipy}, each 
measurement is a $361$-D vector (non-negative entries; sum to $1$). Using the square root parameterization \cite{brody1998statistical,srivastava2007riemannian}, the data at each voxel lies on the positive part of $\mathbf{S}^{361}$ manifold. 

We seek to generate a 3D brain image where each voxel is a ODF from
the corresponding DTI image (each voxel is a $3\times 3$ $\textsf{SPD}$ matrix). To make the setup more challenging (and scientifically interesting), we generate the DTI images only from randomly under-sampled diffusion directions. We now explain the \begin{inparaenum}[\bfseries (a)] \item rationale for the application
\item data description \item model setup \item evaluations. \end{inparaenum}
{\it Note that in the experiment, since we draw samples from the distribution on the latent space, conditioned on DTI, to get the target representation, we call it generation rather than reconstruction.}

\vspace*{0.1cm}
\noindent
{\bf Why generating ODF from DTI is important?} For dMRI, different types of acquisitions involve longer/shorter acquisition times.
Higher spatial resolution images (e.g., ODF) involves a longer acquisition time
($7$--$12$ mins per scan versus $35$ mins for an ODF multi-shell scan) and this is problematic, especially for children and the elderly. To shorten the acquisition time with minimal compromise in the image quality, we require mechanisms to transform data acquired from shorter acquisitions (DTI) to a higher spatial resolution image: a field (or image) of ODFs. {\it This serves as our main motivation.}

However,  
\begin{inparaenum}[\bfseries (a)] \item the per voxel degrees of freedom for ODF representation is $361$ (lies on $\mathbf{S}^{361}$) while for DTI is $6$ (lies on $\textsf{SPD}(3)$). Hence, it is an ill-posed problem. \item requires mathematical tools to ``transform'' from one manifold (DTI representation) to another (ODF representation) while preserving structural information. \end{inparaenum}  Now, we describe some details of the data, models and present the results.

\begin{table}[!b]
\begin{center}
  \scalebox{0.7}{
\centering
\begin{tabular}{c|cccc|cc} 
\topline\myrowcolour
{\bf Dataset} & \multicolumn{4}{c|}{{\bf Age}} & \multicolumn{2}{c}{{\bf Gender}} \\
 \myrowcolour
& 22-25 & 26-30 & 31-35 & 36+ & Female & Male\\
 \midtopline
All & 224 & 467 & 364 & 10 & 575(54.0\%) & 490(46.0\%)\\
Train & 178 & 370 & 295 & 9 & 463(54.3\%) & 389(45.7\%)\\
Test & 46 & 97 & 69 & 1 & 112(52.6\%) & 101(47.4\%)\\
\bottomline
\end{tabular}
}
\end{center}
\caption{The demographics used in the study.}
\label{stat_connectome}
\end{table}

\vspace*{0.1cm}
\noindent
{\bf Dataset.}
The dataset for our method is the Human Connectome Project (HCP) \cite{van2013wu}. The total number of subjects  with diffusion measurements available is $1065$: $852$ were used as training and $213$ as the test set. Demographic details are reported in Table \ref{stat_connectome} (please see \cite{van2013wu} for more details of the dataset). All raw dMRI images are pre-processed with the HCP diffusion pipeline with FSL's `eddy' \cite{andersson2016integrated}. After correction, ODF and DTI pairs were obtained using the Diffusion Imaging in Python (DIPY) toolbox \cite{garyfallidis2014dipy}. 
Due to the memory requirements of the model and 3D nature of medical data,
generation of an ODF image of the entire brain at once remains out of reach at this point,
hence we resize the original data into $32 \times 32 \times 32$ but the process can proceed in a sliding window fashion as well.

\begin{figure}[!t]
        \centering
                \includegraphics[width=0.99\columnwidth]{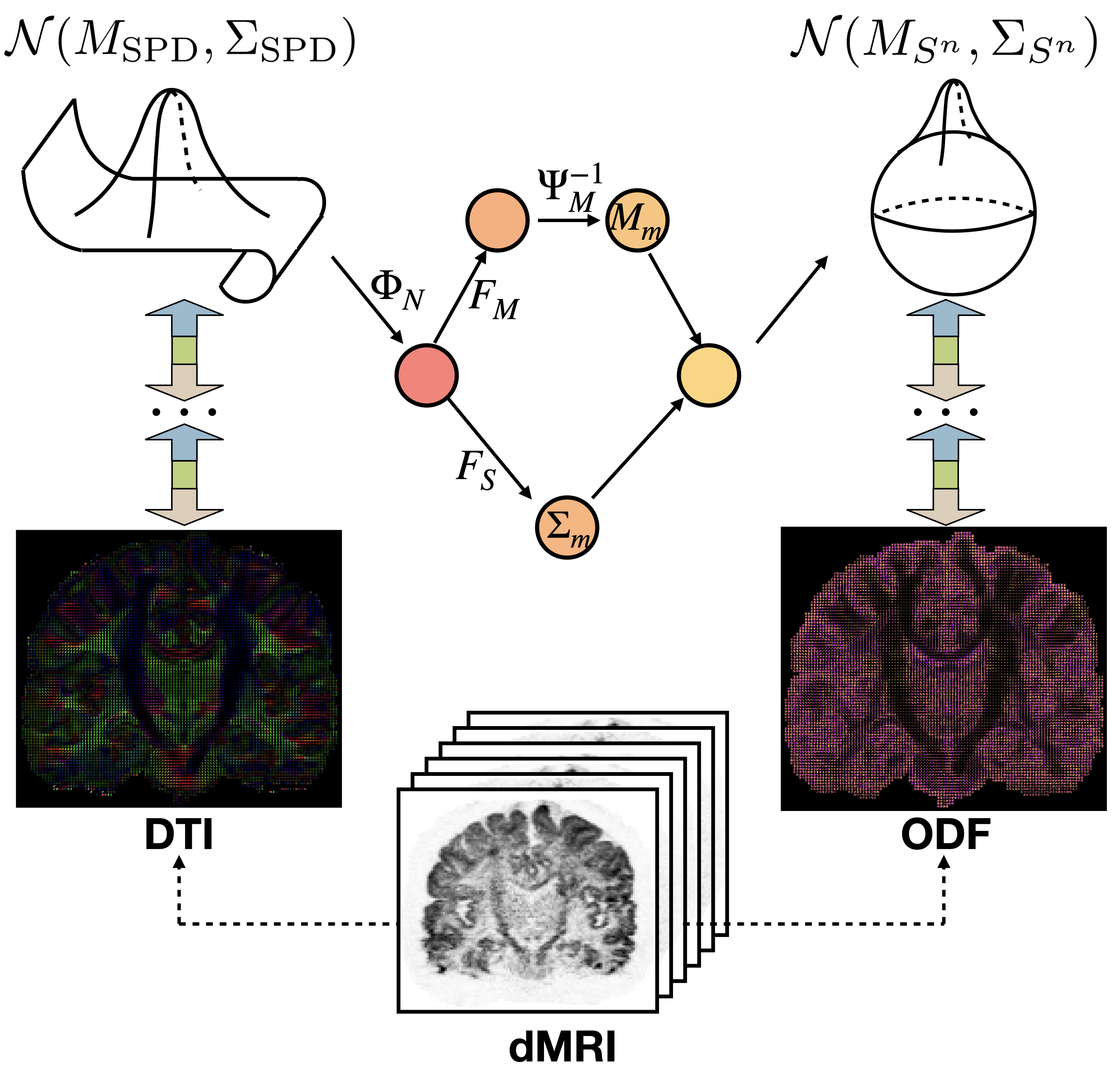}
               \caption{The transformation from DTI to ODF. Both are generated from dMRI. But there might not be dMRI available in some situations. Thus, we want to train the network to transfer DTI to ODF. The latent space is the Gaussian distribution variable.}\label{dual_glow}
\end{figure}

\begin{figure*}[!htb]
        \centering
        \includegraphics[width=0.95\textwidth]{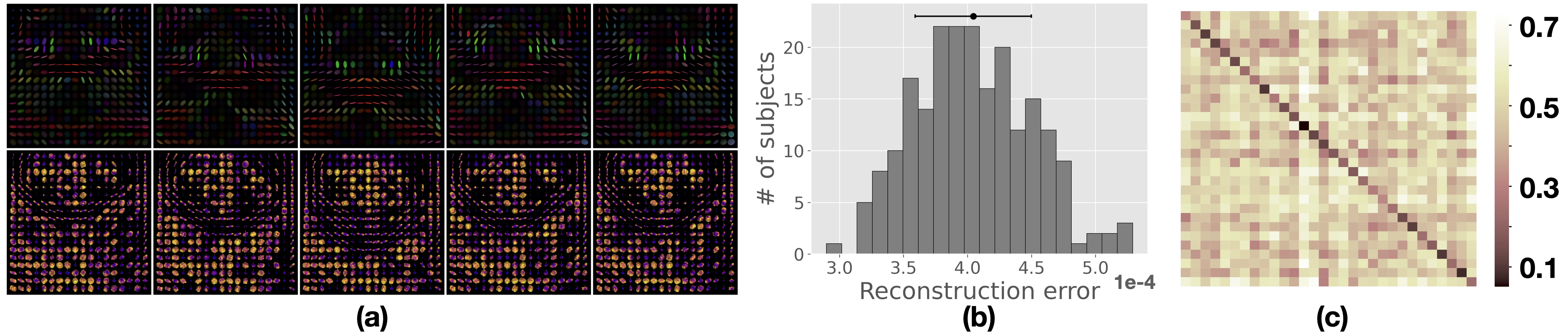}
               \caption{
               (a) Generated ODF from corresponding DTI. Each pair here contains the input DTI {\it (top)} and the generated ODF {\it (bottom)}
               (b) The distribution of reconstruction error over the testing population.
               (c) Reconstruction error over the test population shows that the generated image is closest to its own corresponding image (diagonal dominance))}\label{error_plot}
\end{figure*}

\vspace*{0.1cm}
\noindent
{\bf Reduction in the memory costs.}
Since the entire 3D models for brain images are still too large to fit into the GPU memory, we need to further simplify the model without sacrificing the performance too much. Recently,  
NanoFlow \cite{lee2020nanoflow} was introduced to reduce the number of parameters for sequential data processing. The assumption of NanoFlow is that the Affine Coupling layer, if fully trained, can estimate the distribution for any parts of the input data in a fixed order. There will be some performance drop compared with training different Affine Coupling layers for different parts of the data. But the gain from reducing the parameters is significant. Thus, in our setup, due to the large 3D input, we apply the NanoFlow trick for DTI and ODF separately. 
For example, for the DTI data, we first split the entire data into $2\tau$ slices called $\{X_1, X_2, ..., X_{2\tau}\},\tau\geq 1$. Then we can share the two neural networks $S$ and $T$ in the Affine Coupling layer among these slices. The input of two neural networks $S$ and $T$ in Affine Coupling layer would be $X_{2k-1}, k=1,2,...,\tau$, while the output will be the estimated mean and variance of $X_{2k}, k=1,2,...,\tau$ respectively. Due to sharing weights, the number of parameters reduces and becomes feasible for training our 3D DTI and ODF setups. 

\vspace*{0.1cm}
\noindent
{\bf Model Setup.} 
In order to set up our model, we first build two flow-based streams for DTI and ODF separately. Then, 
in the latent space, we train a transformation operating between the Gaussian distribution variable on the manifold $\mathbf{S}^{361}$ and the Gaussian distribution variable on the manifold $\textsf{SPD}(3)$. 
This architecture with two flow-based models and the transformation module 
can be jointly trained as shown in Fig. \ref{dual_glow}. 
We use $6$ basic blocks of our manifold GLOW, and after every $2$ blocks, reduce the resolution by half. This 
setup is the same for both DTI and ODF. 
We use $3$ residual network blocks to map the 
latent space from DTI to ODF. The samples are presented 
to the model in paired form, i.e., a DTI image (field of SPD matrices) and a corresponding ODF image (a field of ODFs). 
To reduce the number of parameters for this 3D data, we use a similar idea as NanoFlow \cite{lee2020nanoflow} that shares the Affine Coupling layer for DTI and ODF separately, with setting $\tau=32$. 
As a comparison, for the baseline model $\mathcal{M}_e$-flow, the learned dimension will be $32\times 32\times 32 \times d$ where $d=6$ for DTI and $d=361$ for ODF. While 
$\mathcal{M}_e$-flow could be trained for our texture 
experiments, here, the memory requirements are 
quite large, quantitatively the number of parameters required for $\mathcal{M}_e$-flow and our model are $1.3e18$ and $2.1e8$ respectively. 
A similar situation arises in the Euclidean space version of GLOW which also does not leverage the intrinsic Riemmanian metric: therefore, the memory cost will be $6000\times$ more than the natural images which have dimension $224\times 224 \times 3$. This is infeasible even on clusters and therefore, results from these baselines are 
very difficult to obtain.

\vspace*{0.1cm}
\noindent
{\bf Choice of metrics.} We will use ``{\it reconstruction error}'' using the distance in Table \ref{manifoldExample}.
Although the task here is generation, measuring reconstruction error assesses how ``similar'' the original ODF is to the generated ODF,
generated directly from the corresponding DTI representation.
We also perform a group difference analysis to identify statistically different regions across groups (grouped by a dichotomous variable).
Since HCP only includes healthy subjects (HCP aging is smaller), we can perform a group difference test
based on gender, i.e., {\it male vs. female}. We evaluate overlap: how/whether group-wise different regions on the generated/reconstructed
data agrees with those on the actual ODF images. 

\begin{figure}[!htb]
        \centering
        \includegraphics[width=0.9\columnwidth]{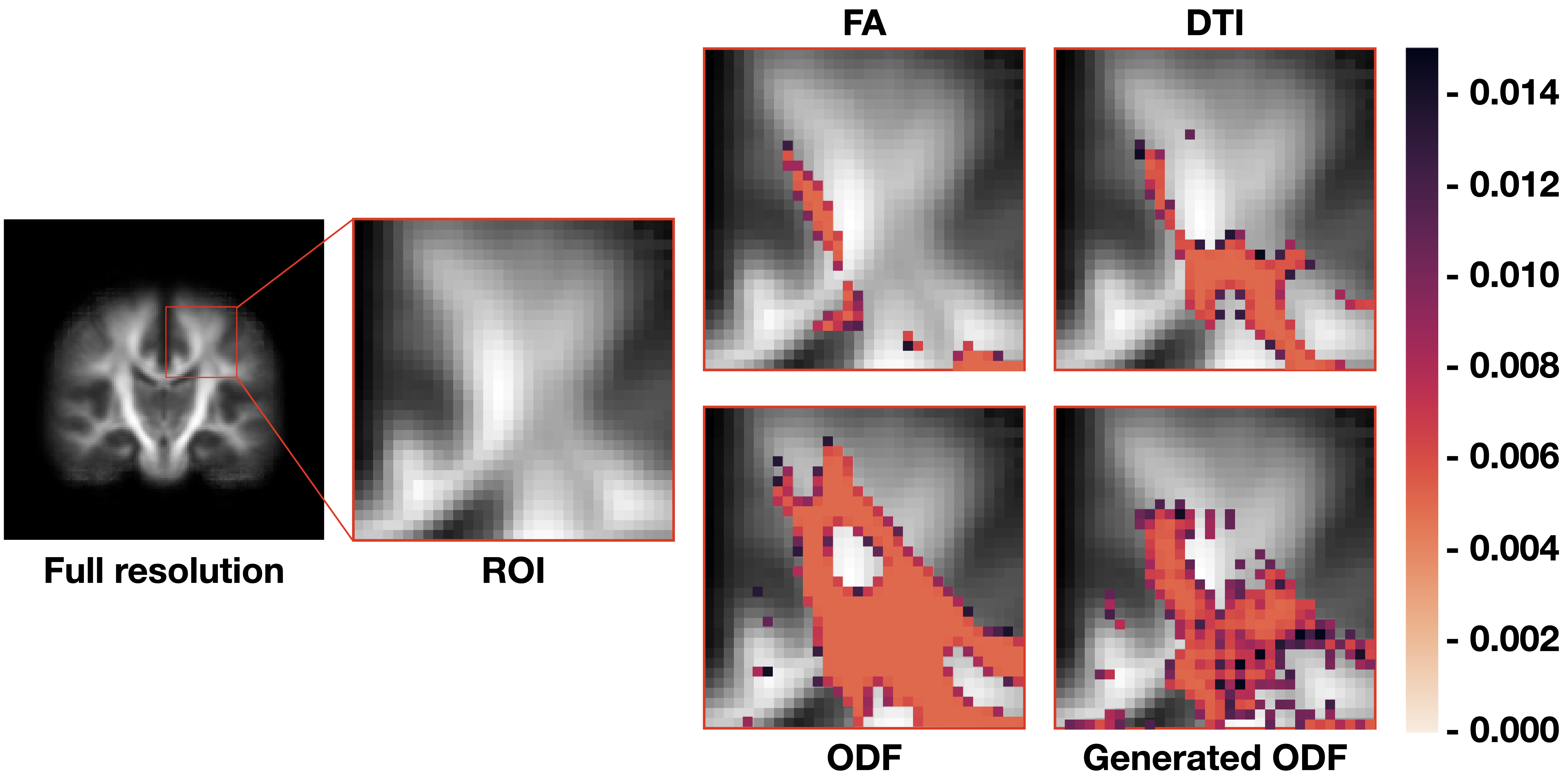}
               \caption{The $p$-value of one of the ROIs of the entire brain scan with full-resolution. We show that our proposed method can generate meaningful ODF with respect to the group level differences.
               }\label{p_value_heatmap}
\end{figure}

\vspace*{0.1cm}
\noindent
{\bf Generation results.} 
We present quantitative and qualitative results for generation of ODF from its DTI representation. In Fig. \ref{error_plot}(a), we show a few example slices from the given DTI and the generated ODF. 
Overall, the reconstruction error was $4.0(\pm 0.45)\times 1e\text{-}4$.  
Since perceptually comparing fidelity between generated 
and ground truth images is difficult, we perform the following quantitative analysis: 
\begin{inparaenum}[\bfseries (a)]  
\item a histogram of the reconstruction error over all $213$ test subjects (shown in Fig. \ref{error_plot}(b)) 
\item an error matrix showing how similar the generated ODF image 
is with the other ``incorrect'' samples of the population. 
The goal is to assess if the generated ODF is distinctive across different samples (shown in Fig. \ref{error_plot}(c)) 
\end{inparaenum}.  From the histogram presented in Fig. \ref{error_plot}(b), 
we can see that the reconstruction error is consistently low
over the entire test population. Now, we generate Fig. \ref{error_plot}(c) as follows. For each subject in the test population, we randomly select $29$ samples (subjects) from the population and compute the reconstruction error with the generated ODF. 
This gives us a $30\times 30$ matrix (similar to the confusion matrix).
Fig. \ref{error_plot}(c) shows the average of $10$ runs: lighter shades mean a larger reconstruction error.
So, we should ideally see a dark diagonal, which is approximately depicted in the plot. 
This suggests that for the test population, the generation is meaningful (preserves structures) and 
distinctive (maintains variability across subjects). 
There are only few experiments described in the literature on generation of dMRI data \cite{huang2019manifold,anctil2020manifold}.
While \cite{huang2019manifold} shows the ability to generate 2D ($32\times 32$) DTI, the
techniques described here can operate on 3D ODF ($\mathbf{S}^{361}$) data and should offer improvements. 

\vspace*{0.1cm}
\noindent
{\bf Group difference analysis.}
We now quantitatively measure
if the reconstruction is good enough so that the generated samples can be a good proxy for downstream statistical analysis and yield improvements 
over the same analysis performed on DTI. 
We run permutation testing with $10000$ independent runs and compute the per-voxel $p$-value to see which voxels were statistically
different between the groups for
the following settings \begin{inparaenum}[\bfseries (a)] \item original ODF \item generated ODF \item DTI \item functional anisotropy (FA) representation (commonly used summary of DTI). \end{inparaenum} Both DTI and FA are commonly
used for assessing statistically significant differences across genders \cite{menzler2011men,kanaan2012gender}. But since
ODF contains more structural information than either the FA or DTI, our generated ODF should be able to
pick up more statistically significant regions over DTI or FA. We evaluate the intersection of significant regions with the original ODF (the original ODF contains the most information). We compute the {\it intersection over union} (IoU) measure. For the whole brain, FA will have IoU 0.04, while DTI has IoU 0.16. The generated ODF has IoU 0.22. 
We see that the generated ODF has a larger intersection in the statistically significant regions with
the original ODF and offers improvements over DTI. This provides
some evidence that the generated ODF preserves the signal that is different across the male/female groups. 
We also show a zoomed in example of a ROI for the full-resolution images in Fig. \ref{p_value_heatmap}. The $p$-values for different ROIs are all $< 0.001$ in both the original ODF and our generated ODF, indicating consistency of our results, at least in terms of regions identified in downstream statistical analysis. Note that the analysis on the real ODF images serves as the ground truth.

\section{Conclusions}
A number of deep neural network
formulations have been extended to manifold-valued data in the last two years. 
While most of these developments are based on models such as
CNNs or RNNs, in this work, we study the generative regime: we
introduce a flow-based generative model on the Riemannian manifold.
We show that the
three types of layers, Actnorm, Invertible $1\times 1$ convolution, and Affine Coupling layers in such models,
can be generalized/ adapted for manifold-valued data in a way that preserves invertibility.
We also show that with the transformation in the latent space between the two manifolds,
we can generate manifold-valued data based on the information from another manifold. 
We demonstrate good generation results in the representation of ODF given DTI on the Human Connectome dataset.
While the current formulation shows mathematical feasibility and promising results, additional work on the methodological
and the implementation side is needed to reduce the runtime to a level where the tools can be deployed in
scientific labs.

\section{Acknowledgements}
This research was supported in part by grant  1RF1AG059312-01A1 and NSF CAREER RI \#1252725.

{
\bibliography{citations}
}

\end{document}